\pdfoutput=1
\documentclass{openmoss}

\usepackage{pifont}        % \ding: check / cross marks
\usepackage{enumitem}      % list styling
\usepackage{float}         % [H] float placement
\usepackage{makecell}      % multi-line table cells
\usepackage{tabularx}      % auto-width columns
\usepackage{multirow}      % multi-row table cells
\usepackage{xcolor,colortbl} % \rowcolor for highlighting ``ours''
\usepackage{algorithm}        % algorithm float
\usepackage{algpseudocode}    % pseudocode
\usepackage{calc}             % \widthof, for aligning the dialogue sample (\S2.3)
\usepackage{titlesec}         % override run-in \paragraph headers to bold (below)
\crefname{algorithm}{Algorithm}{Algorithms}
\Crefname{algorithm}{Algorithm}{Algorithms}

\titleformat*{\paragraph}{\bfseries}

\graphicspath{{assets/}{./}}

\definecolor{oursgray}{gray}{0.95}     % shade the ``our method'' row
\definecolor{tickG}{HTML}{00C853}
\definecolor{crossR}{HTML}{FF1744}

\newcommand{\bnum}[1]{\textbf{#1}}     % bold a best value
\newcommand{\cmark}{\textcolor{tickG}{\bfseries\ding{52}}}
\newcommand{\xmark}{\textcolor{crossR}{\bfseries\ding{56}}}

\definecolor{tokvid}{HTML}{1565C0}     % <|video|>   : a frame arrives
\definecolor{toksil}{HTML}{C2410C}     % <|silence|> : the model stays silent
\newcommand{\vtok}{\textcolor{tokvid}{<|video|>}}
\newcommand{\stok}{\textcolor{toksil}{<|silence|>}}
\newcommand{\rolerow}[1]{\{"role": \makebox[\widthof{"assistant", }][l]{"#1",} "content": }

\newtcolorbox{promptbox}[2][]{
  colback=white, coltext=black,
  arc=3mm, boxrule=0.5pt, colframe=black!60!white,
  title={#2}, colbacktitle=black, coltitle=white, fonttitle=\bfseries,
  top=8pt, bottom=8pt, left=10pt, right=10pt,
  breakable,
  before upper={\linespread{1}\selectfont
    \setlength{\parskip}{1ex plus 0.2ex minus 0.2ex}\setlength{\parindent}{0pt}},
  #1
}

\title{MOSS-Video-Preview: Toward Real-Time Video Understanding via Cross-Attention}

\author[1*\dagger]{Pengyu Wang}
\author[1\dagger]{Chenkun Tan}
\author[1\dagger]{Shaojun Zhou}
\author[\dagger]{Wei Huang}
\author[1\dagger]{Qirui Zhou}
\author[1\dagger]{Zhan Huang}
\author[1\dagger]{Zhen Ye}
\author[1,2\dagger]{Jijun Cheng}
\author[2]{Xiaomeng Qian}
\author[1]{Yanxin Chen}
\author[2]{Xingyang He}
\author[1,2]{Huazheng Zeng}
\author[1]{Chenghao Wang}
\author[1]{Pengfei Wang}
\author[1]{Hongkai Wang}
\author[1]{Shanqing Gao}
\author[2]{Yixian Tian}
\author{Chenghao Liu}
\author[1]{Xinghao Wang}
\author[1,2]{Botian Jiang}
\author[1,2\ddagger]{Xipeng Qiu}
\affil[1]{Fudan University}
\affil[2]{Shanghai Innovation Institute}
\authormark{$^{*}$Project Leader\quad $^{\dagger}$Core Contributor\quad $^{\ddagger}$Corresponding Author}

\abstract{% Abstract body. Included by \abstract{\input{../chapters-en/00-abstract}} in main.tex.
Video understanding is shifting from the offline paradigm---taking a fully recorded video as input and producing a single answer after it ends---toward \textbf{real-time interaction}, in which the model perceives new frames while still replying, revises its answer as new evidence appears, and remains silent when there is nothing to say. We present MOSS-Video-Preview to validate this paradigm. Our central claim is that \textbf{perception must not be blocked by generation}; its natural realization is a two-channel architecture. We argue that a cross-attention backbone is better suited to real-time vision--language fusion than the prevailing decoder-only design: visual features enter through a side channel rather than joining the autoregressive sequence, so perception and generation run on separate, non-blocking pathways---reducing the frequency of visual processing and exposing a clean channel-wise interface for independent compression. We complement this architecture with a data synthesis pipeline that converts dense captions into real-time understanding QA whose answers are revised to match what the model has perceived so far, and we then specialize an offline model on these data to elicit real-time behavior. As a preview, we prioritize feasibility over state-of-the-art performance. Our model still trails the strong Qwen2.5-VL-7B baseline overall---a gap we attribute primarily to data and scale rather than the architecture---yet attains competitive offline video and multimodal understanding, remains robust on the spatial and fine-grained temporal reasoning central to real-time use, and acquires behaviors that offline models lack: continuous perception, answer revision, and timely silence. On a single H200 with 256 frames per video, it achieves approximately a 5$\times$ speedup in time to first token and 2.7$\times$ higher decoding throughput despite its larger size, with negligible degradation in offline ability. Taken together, our study of paradigm, architecture, and data outlines a viable path toward real-time video understanding.
}

\checkdata[GitHub]{\url{https://github.com/OpenMOSS/MOSS-Video-Preview}}
\checkdata[Hugging Face]{\url{https://huggingface.co/collections/OpenMOSS-Team/moss-video-preview}}

\begin{document}
\maketitle

\section{Introduction}
\label{sec:intro}

Multimodal models have progressed from question answering on single images~\citep{liu2023llava} to understanding videos that span minutes to hours~\citep{zhang2024llavavideo,wu2024longvideobench}. Most of these models, however, still assume that the video has been fully recorded and is available before generation begins: the model watches the clip through, then answers. This assumption breaks down in situated applications such as smart glasses, embodied robots, livestream assistants, and co-watching agents, where the video is an environmental stream unfolding in the present and new frames keep arriving even while the model is responding. Recent streaming systems~\citep{chen2024videollmonline,wang2025mmduet,qian2025dispider} move beyond this offline setting by responding while the video is still playing, but they still treat perceiving and replying as separate phases. What these applications ultimately demand is \textbf{real-time interaction}: a model that keeps perceiving while it generates, behaving like an always-on observer---watching when no question has been asked, speaking up when a salient event occurs, revising its answer when the situation changes, and waiting quietly when there is nothing to say.

The term ``real-time'', however, is often conflated with ``streaming'' in the literature, and the boundary between the two capabilities is rarely made explicit. We make this distinction precise; its essence reduces to a single dividing line: \textbf{whether the model can continue perceiving while it is generating a reply}. Offline systems answer only after the entire clip has ended; existing streaming systems can respond along the timeline, but \textbf{stop ingesting new frames during the generation of a reply} and therefore cannot react in time to changes that occur within that reply window---any correction is delayed by at least one reply length. Real-time systems, by contrast, require perception and generation to be \textbf{concurrent}: to keep observing while answering, to revise or even interrupt the current reply the moment evidence appears, and to use an explicit silence decision to avoid re-describing a static scene. The distinction is not whether an answer can be revised at all---streaming systems too can revise on a later turn---but \textbf{whether the revision is timely}; the root cause is precisely the constraint that defines real-time interaction: \textbf{perception must not be blocked by generation}.

This constraint shapes our architectural choice. For continuous frame reading and continuous token generation to coexist without interference, the natural realization is a two-channel architecture. Decoder-only designs~\citep{bai2025qwen3vl,an2026llavaonevision2} are not incapable of supporting real-time interaction---one can keep inserting frames into the token stream---but we argue that a cross-attention backbone~\citep{alayrac2022flamingo} fits the real-time requirement more naturally: visual features are injected as a side channel without joining the autoregressive sequence, so perception and generation are physically separated into two pathways, which further brings a lower frequency of visual processing and a cleaner channel-wise interface for independent compression. The architecture grants the model the \textbf{ability} to look and answer in tandem, but the \textbf{behavior} of when to revise and when to remain silent must be learned from data---and today's static caption and QA corpora contain neither trajectories in which answers evolve with the stream nor supervision for silence. We therefore introduce a real-time data synthesis pipeline and instantiate it as MOSS-Video-Preview.

We state up front what this work is meant to be: \textbf{a preview in the spirit of a position paper}. The question is whether the real-time video understanding paradigm and a cross-attention backbone are feasible and effective---not whether the model reaches state-of-the-art. Its emphasis is therefore on validating the paradigm and architecture rather than on scale or completeness: data scaling, exhaustive ablations, and a quantitative protocol for real-time understanding---which the field still lacks---lie beyond the scope of this preview.

This work makes the following contributions:
\begin{itemize}[leftmargin=1.4em, itemsep=2pt, topsep=2pt]
  \item \textbf{The real-time video understanding paradigm and its formalization.} We make precise the distinction between offline, streaming, and real-time paradigms---its essence being whether the model can continue perceiving during reply generation---formalize real-time behavior as an interleaved sequence of text and frames with a \texttt{<|silence|>} token, and identify the key gap in evaluation: decision-level latency goes unmeasured, and accuracy alone can be inflated by ``stalling the answer''.
  \item \textbf{A cross-attention backbone tailored to real-time interaction.} We argue that this fit is \textbf{structural}. First, the two pathways are physically separate, so perception never blocks generation and the model can look and answer in tandem. Second, the same separation lowers inference cost: visual content is retrieved in only a few layers, and the two pathways can be compressed independently. We instantiate the design on Llama-3.2-11B-Vision~\citep{grattafiori2024llama3}, with adaptations including per-frame temporal positional encoding and 2D pooling compression.
  \item \textbf{Real-time data synthesis and basic data curation.} We propose a real-time data synthesis pipeline that converts static understanding data into supervision in which ``the instruction stays fixed, the best reply evolves with the stream, and silence fills the rest''. Alongside, we curate large-scale basic-understanding and instruction data to first train a strong offline model, then specialize it into a real-time one.
  \item \textbf{Real-time training and inference.} We inject real-time behavior into a strong offline model with a real-time SFT stage---a mixed corpus of real-time and offline QA, with two system prompts distinguishing the two modes---and realize it as a silence-gated two-state loop: a per-step threshold gates ``answer or stay silent now'', and a single set of weights exposes both offline and real-time entry points.
\end{itemize}

The experiments validate the design from three angles. The model attains competitive offline video and multimodal understanding: it still trails the strong baseline Qwen2.5-VL-7B overall on general benchmarks, but performs robustly on the spatial and fine-grained temporal dimensions that matter most for real-time understanding. On a single H200 with 256 frames sampled per video, it achieves about a 5$\times$ speedup in time to first token and a 2.7$\times$ improvement in decoding throughput, and this advantage stems from the architecture, not an engineered serving stack. Finally, specializing the model from offline to real-time incurs negligible degradation in its offline understanding. Together, these results support this work's position---that the real-time video understanding paradigm and a cross-attention backbone form a viable and effective starting point.

\section{The Real-Time Paradigm}
\label{sec:problem}

``Streaming'' is applied to a broad range of systems, and the literature rarely states precisely what such a model can and cannot do. This section makes the distinction explicit: we separate the \emph{real-time} paradigm targeted in this work from the offline paradigm and from existing streaming systems, and give its definition, its constraints, and the difficulty of evaluating it.

\subsection{Offline, streaming, and real-time}
\label{subsec:paradigms}

The \emph{essential} difference is whether \emph{the model can continue perceiving while it is generating a reply}:
\begin{itemize}[leftmargin=1.4em, itemsep=2pt, topsep=2pt]
  \item \textbf{Offline.} The model takes the entire clip and produces a single answer; by the time it answers, the video has already ended.
  \item \textbf{Streaming (existing).} Frames arrive in temporal order, and the model may respond mid-playback; \emph{but it stops ingesting new frames during the generation of a reply}. It therefore cannot react in real time to changes that take place inside that reply window---any correction must wait until the current reply ends and the model returns to a perception state, so the correction is \emph{delayed} by one reply length.
  \item \textbf{Real-time (ours).} The model \emph{keeps perceiving new frames while it is replying}, and can therefore revise---or even interrupt---its current reply the moment evidence appears.
\end{itemize}

\begin{figure}[t]
  \centering
  \includegraphics[width=\linewidth]{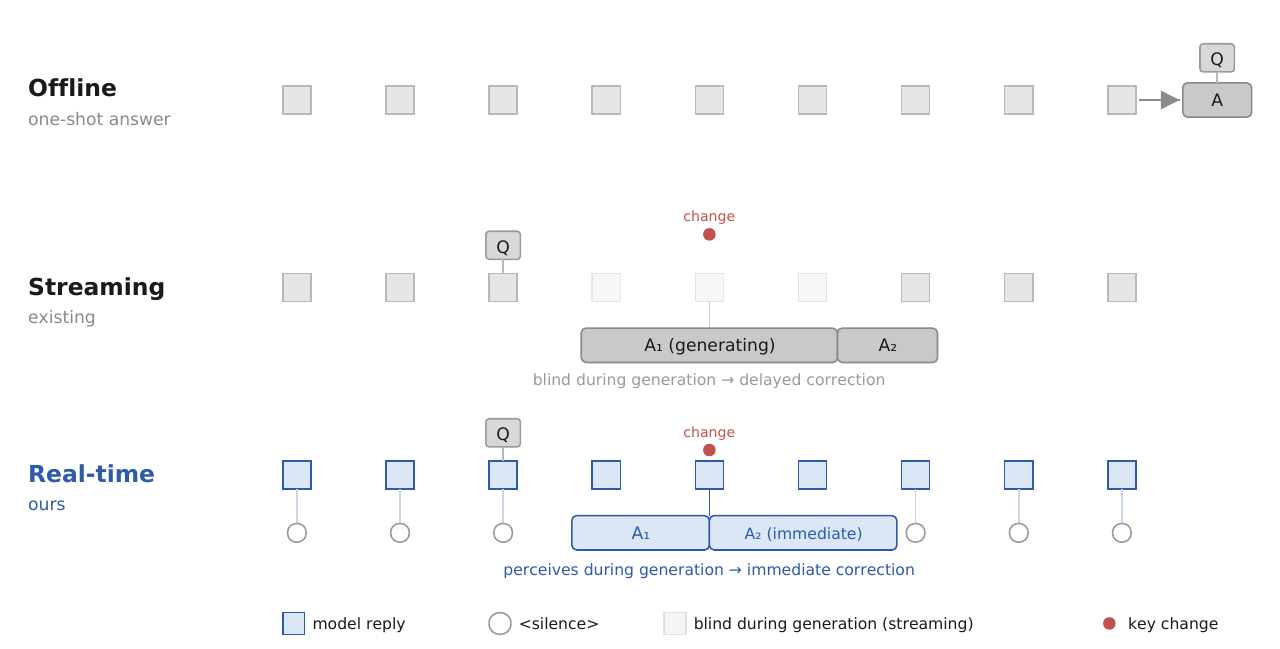}
  \caption{Three paradigms of video understanding. \textbf{Offline}: answers in one shot after the clip has ended. \textbf{Streaming}: can respond on time, but \emph{stops perceiving while a reply is being generated}; any change inside that window must wait until the current reply ends to be addressed on the next turn (delayed). \textbf{Real-time} (ours): \emph{keeps perceiving during reply generation}, so a revision can happen the moment a relevant change appears.}
  \label{fig:paradigms}
\end{figure}

We emphasize: the difference is \emph{not} whether the answer can be revised at all (streaming too can revise on a later turn) but \emph{whether the revision is timely}. The root cause is that streaming makes perception and generation serial and mutually exclusive, whereas real-time lets them run concurrently. \Cref{fig:paradigms} illustrates the three paradigms, and \Cref{tab:paradigms} summarizes the capability differences.

\begin{table}[b]
  \centering
  \caption{Capabilities of the three video-understanding paradigms.}
  \label{tab:paradigms}
  \setlength{\tabcolsep}{8pt}
  \renewcommand{\arraystretch}{1.15}
  \begin{tabular}{@{}lccc@{}}
    \toprule
    \textbf{Property} & \textbf{Offline} & \textbf{Streaming} & \textbf{Real-time (ours)} \\
    \midrule
    Frames arrive in temporal order                       & \xmark  & \cmark                  & \cmark    \\
    \textbf{Keeps perceiving during reply generation}     & ---     & \xmark                  & \cmark    \\
    Time-to-correction                                    & ---     & deferred to next turn   & immediate \\
    Proactive silence when nothing to say                 & \xmark  & partial                 & \cmark    \\
    \bottomrule
  \end{tabular}
\end{table}

\subsection{Constraints}
\label{subsec:constraints}

There is only one core constraint; the rest organize around it.
\begin{itemize}[leftmargin=1.4em, itemsep=2pt, topsep=2pt]
  \item \textbf{(Core) Perception must not be blocked by generation.} The model keeps ingesting frames while producing a reply---this is what separates real-time from streaming and enables ``timely correction''.
  \item \textbf{Continuous perception.} The model keeps observing even when no question has been asked, because salient events often occur before or between questions.
  \item \textbf{Low latency.} The per-frame delay from arrival to ``answer or silence'' must be low, otherwise the interaction loop cannot close (this is the system-level latency, distinct from the decision-level one defined below).
  \item \textbf{Silence decision.} The model must explicitly decide ``should I speak now?''; continuous perception without a silence mechanism leads to the same static scene being described over and over.
\end{itemize}

\subsection{Formalization}
\label{subsec:formalization}

We model a real-time interaction as a sequence of text and video frames interleaved in a multi-turn dialogue. Denote the interaction as $x_{1:N}$, where each $x_i$ is drawn from the text vocabulary, from the video-frame placeholder \texttt{<|video|>}, or from the silence symbol \texttt{<|silence|>}. Video frames are inserted into the sequence at their arrival times as exogenous inputs (their visual features are injected via cross-attention); at every step the model autoregressively predicts the next text or \texttt{<|silence|>} token. The dialogue structure (a \texttt{role}/\texttt{content} sequence) of one training sample is illustrated below (the system prompt is abbreviated):

\begin{tcolorbox}[
    enhanced,
    colback=gray!4, colframe=black!55, boxrule=0.6pt, arc=2mm,
    left=10pt, right=10pt, top=6pt, bottom=6pt,
    title={\ttfamily\bfseries\footnotesize role / content sequence of one training sample},
    coltitle=white, colbacktitle=black!62, fonttitle=\footnotesize,
    toptitle=2pt, bottomtitle=2pt]
\footnotesize\ttfamily\setlength{\parindent}{0pt}\linespread{1.0}\selectfont\raggedright
[\\
\hspace*{1.2em}\rolerow{system}"You perceive the environment in real time and interact with the user. Keep observing even when not asked; promptly update your answer when relevant changes occur; output \stok\ when there is nothing to say."\},\\
\hspace*{1.2em}\rolerow{user}""\},\\
\hspace*{1.2em}\rolerow{assistant}"\stok\vtok\stok"\},\\
\hspace*{1.2em}\rolerow{user}"What is happening here?"\},\\
\hspace*{1.2em}\rolerow{assistant}"The entrance gate\vtok\ stands quietly, with 'Beijing Zoo'\vtok\ shown in black on top. For now\vtok\ there is no notable activity.\vtok\stok\vtok\stok\vtok\stok\vtok\ A keeper now walks up and unlocks the gate.\vtok\stok\vtok\stok"\}\\
]
\end{tcolorbox}

Three features of this representation map directly onto the paradigm distinctions above. \emph{First}, the interaction starts from a default turn in which the user input is empty and the assistant has already begun receiving frames and answering with \texttt{<|silence|>}---this captures the always-on perception even when no question is being asked (the continuous-perception constraint above). \emph{Second}, and most crucially, video frames are interleaved \emph{inside} the assistant's reply: the model keeps receiving new frames while it is generating an answer, so perception and generation are concurrent on the same sequence. This is the essence that separates real-time from streaming. \emph{Third}, \texttt{<|silence|>} explicitly encodes ``no reply at this moment''. After an answer is complete, every subsequent frame is met with \texttt{<|silence|>}, indicating that the model is still observing but has nothing new to report. When a question has been raised but the relevant evidence has not yet appeared, the model also emits \texttt{<|silence|>} and waits, replying only when the evidence allows. When a relevant change later surfaces during the silence period---such as the keeper approaching and unlocking the gate in the example---the model proactively breaks silence to supplement the answer, a ``timely correction''.

Under this representation the three paradigms are special cases of the same sequence model: offline produces a single block of output only after the frame stream ends; streaming admits no new frame during the generation of a reply; and real-time both allows frames to keep interleaving inside the reply and models \texttt{<|silence|>} explicitly. The training supervision for \texttt{<|silence|>} is discussed later.

\subsection{The evaluation dilemma}
\label{subsec:eval-gap}

A real-time evaluation must \emph{jointly} measure ``accuracy'' and ``timeliness'', and timeliness itself has two flavors: \textbf{system-level} (e.g., TTFT, MaxFPS---how fast the inference is) and \textbf{decision-level}~\citep{lin2024streamingbench,li2025ovobench} (the wait between an event ending and the model choosing to reply---how fast the perception-to-judgment is). The two are orthogonal: a model with fast inference may still take a long time to update its answer. More importantly, \emph{looking at accuracy alone can be inflated by ``stalling the answer''}: on temporal-grounding-style tasks, a model that waits until well after the event has ended---when its information is unambiguous---to emit an answer can score artificially high while flouting the very purpose of being real-time.

\emph{This work does not include an evaluation of real-time understanding}, nor does it propose a decision-level latency benchmark---this is an open problem of the paradigm, and is left to future work (which also considers using it as a reinforcement-learning reward, $R = \text{acc} - \lambda \cdot \text{delay}$).

\section{Architecture}
\label{sec:arch}

The starting point of this section is the definition of real-time video understanding established earlier, whose single core demand is that \emph{perception of new frames must continue while a reply is being generated}. To meet it natively at the architectural level, the most natural realization is a \emph{two-channel} architecture: visual perception and language generation run along two pathways that do not block each other. We therefore choose the cross-attention backbone~\citep{alayrac2022flamingo} over the prevailing decoder-only design as our vision--language fusion paradigm: each frame takes only a single \texttt{<|video|>} placeholder in the text sequence, and its hundreds or thousands of visual features do not enter the autoregressive sequence but are exposed as side-channel \textbf{keys and values} that are retrieved by text-side \textbf{queries} in a small number of layers of the backbone. On this backbone we add three adaptations tailored to real-time interaction and long video:
\begin{enumerate}[label=(\arabic*), leftmargin=1.6em, itemsep=2pt, topsep=2pt]
  \item we add a \emph{per-frame rotary positional encoding} to cross-attention, putting each frame onto the same unified temporal position axis as the text;
  \item we apply \emph{2D pooling} before injection to compress the visual tokens of each frame, reducing the K/V volume injected into the backbone;
  \item we wire up an \emph{incremental frame injection} pathway and a streaming inference interface, so that one and the same model can both answer a fully recorded clip offline and look-and-answer at the same time.
\end{enumerate}

We instantiate the design with \textbf{Llama-3.2-11B-Vision}~\citep{grattafiori2024llama3}.

\subsection{Overview}
\label{subsec:arch-overview}

\begin{figure}[t]
  \centering
  \includegraphics[width=\linewidth]{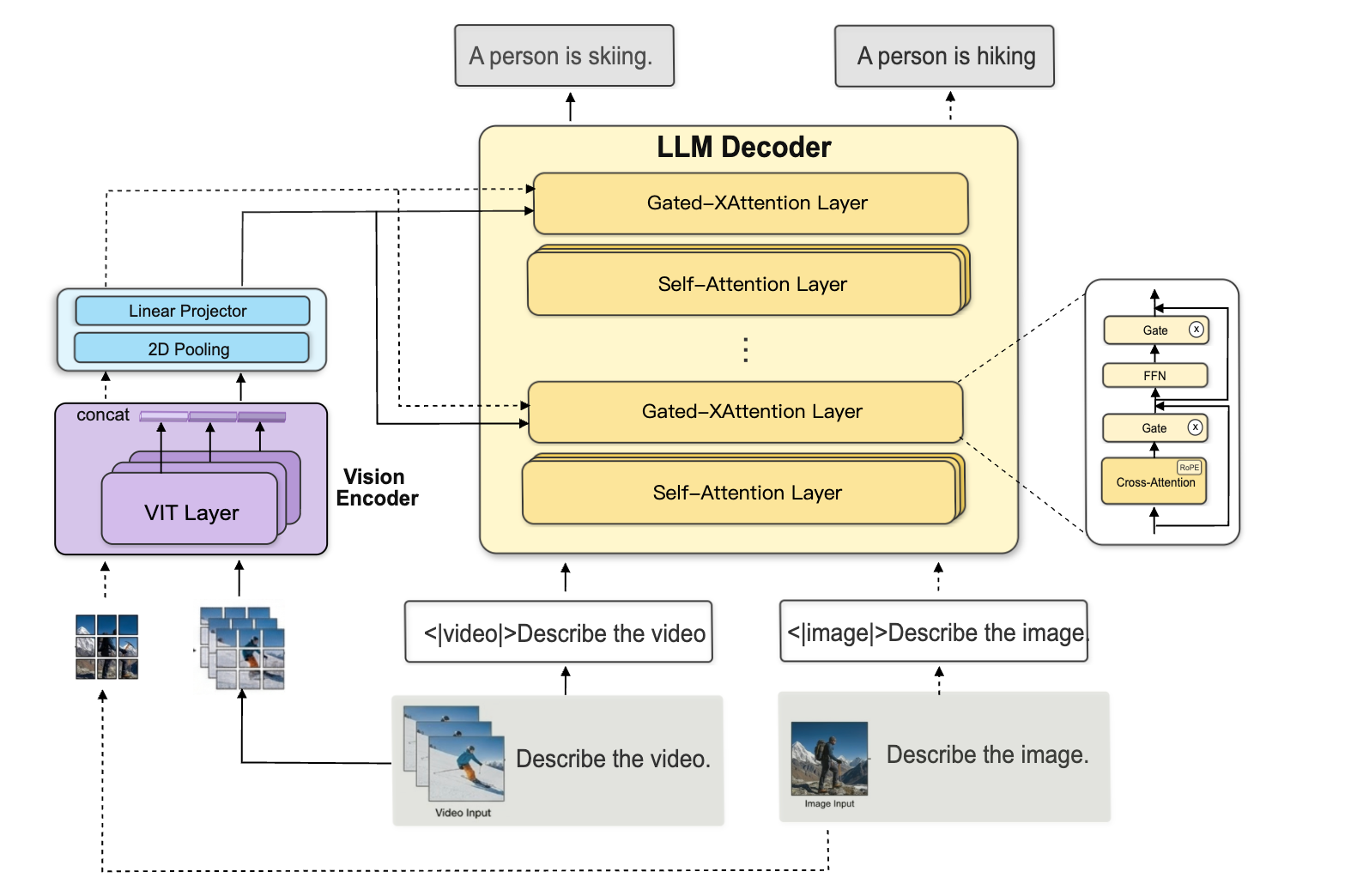}
  \caption{Overall architecture of MOSS-Video-Preview. Each frame or image is encoded by the ViT, spatially compressed by 2D pooling, and projected to the LLM hidden size; the resulting visual features are exposed as keys/values and retrieved by text-side queries at the gated cross-attention layers, while the self-attention layers carry language generation. Right: a gated cross-attention block---RoPE-equipped cross-attention and an FFN, each behind a $\tanh$ gate (\Cref{eq:gated-residual}).}
  \label{fig:architecture}
\end{figure}

The model pairs a visual pathway with a language backbone (\Cref{fig:architecture}). The visual pathway has three pieces---\textbf{vision encoder $\rightarrow$ pooling compression $\rightarrow$ projector}---which together encode each frame into a set of visual features; this set is exposed as K/V and retrieved by text-side queries at the cross-attention layers in the language backbone. The text side is a standard autoregressive decoder: $\mathbf{40}$ layers in total, of which the $\mathbf{8}$ layers $[3, 8, 13, 18, 23, 28, 33, 38]$ are gated cross-attention layers, and the remaining 32 are text-only self-attention layers (hidden size 4096).

Measured directly from the weight tensors, the full model has about $\mathbf{10.7}$\,B parameters (commonly referred to as ``11B''), distributed as in \Cref{tab:params}. The increment for ``introducing vision'' falls mainly on the 8 cross-attention adapter layers ($\approx 1.7$\,B) and the vision encoder ($\approx 0.84$\,B); the remaining $\sim 8$\,B of text backbone keeps the scale and structure of the base model.

\begin{table}[t]
  \centering
  \caption{Parameter distribution of MOSS-Video-Preview, measured directly from the weight tensors.}
  \label{tab:params}
  \setlength{\tabcolsep}{8pt}
  \renewcommand{\arraystretch}{1.15}
  \begin{tabular}{@{}lrl@{}}
    \toprule
    \textbf{Component} & \textbf{Parameters} & \textbf{Notes} \\
    \midrule
    Text self-attention layers (32 layers)  & $\approx 7.0$\,B  & carrier of the base Llama language ability      \\
    Gated cross-attention layers (8 layers) & $\approx 1.7$\,B  & the principal extra cost of introducing vision   \\
    Token embeddings + output head          & $\approx 1.1$\,B  &                                                  \\
    Vision encoder (ViT)                    & $\approx 0.84$\,B &                                                  \\
    Cross-modal projector                   & $\approx 0.03$\,B & a single linear layer                            \\
    \bottomrule
  \end{tabular}
\end{table}

On the input side, a video is first sampled into a sequence of frames at a fixed rate (or uniformly), and each frame corresponds to a \texttt{<|video|>} placeholder token in the text sequence. The key point: \emph{visual features do not enter the autoregressive text sequence; they go through the side channel and are retrieved as K/V, layer by layer.}

\subsection{Rationale for cross-attention}
\label{subsec:arch-why}

The contrast group is the prevailing \textbf{decoder-only} approach, where projected visual tokens \emph{join the text sequence} and undergo self-attention with the text. Both routes can process video, and decoder-only can be made real-time by continually interleaving frames into the token stream; but along the axis of ``look-and-answer at the same time'', the cross-attention fit is structural. Of the five points below, the first three are our core reasons for adopting it; the last two are benefits that follow.

\begin{enumerate}[label=\textbf{(\arabic*)}, leftmargin=1.9em, itemsep=4pt, topsep=2pt]
  \item \textbf{Two channels: perception does not block generation (most fundamental).} Cross-attention puts vision and text on \emph{two pathways}: vision enters from the side channel as K/V while text generates autoregressively, so frame reading and token generation are physically separated and mutually non-blocking. Decoder-only can be made real-time (by inserting frames' visual tokens), but it places both on the \emph{same autoregressive sequence}: every incoming frame extends the sequence being generated, so perception and generation share one context that must be orchestrated explicitly. Cross-attention makes this isolation the default.

  \item \textbf{Incremental frame injection enables ``look-and-answer at the same time''.} A newly arrived frame only needs to \emph{append} its visual K/V to the cache at each cross-attention layer, and the next generation step can already attend to it (a \texttt{vision\_cache\_position} tracks where each frame sits). The same weights and cache thus expose two entry points---\texttt{offline\_generate} (whole video given upfront) and \texttt{real\_time\_generate} (frames arrive over time, tokens stream out, and new questions can be interjected mid-generation)---differing only in whether frames arrive all at once or over time.

  \item \textbf{Lower frequency of visual processing $\rightarrow$ faster inference.} Vision is retrieved in only $\mathbf{8/40}$ layers, and its K/V is encoded \emph{once} at prefill and reused at every decoding step---so a decoding step neither re-encodes vision nor, as in decoder-only, drags a large pool of visual tokens through self-attention at every step. In a single-H200 setup with $256$ frames per video (a non-standardized comparison), our 11B model reaches roughly $\mathbf{5\times}$ TTFT and $\mathbf{2.7\times}$ decoding throughput over Qwen2.5-VL-7B~\citep{bai2025qwen25vl} (decoder-only).

  \item \textbf{Channel separation $\rightarrow$ independent compression.} With the channels separate, compression can target the visual side alone---where, in multi-frame video, visual K/V often dominates the context budget---without touching the text. As a first instance, we already 2D-pool each frame's patch grid before injection, cutting visual K/V to roughly $1/\text{stride}^2$; finer channel-wise compression is a natural follow-up. Decoder-only mixes both token types in one sequence, where such a split is hard to enact.

  \item \textbf{Native support for interleaved images / videos and multi-turn dialogue.} Each frame or image maps to one \texttt{<|video|>} placeholder, and a \texttt{cross\_attention\_mask} controls which text span attends to which vision span. Arbitrary interleavings of images, video segments, and multi-turn dialogue are thus supported natively (our model already handles mixed image-text and video input).
\end{enumerate}

\subsection{Injection and the temporal bridge}
\label{subsec:arch-bridge}

At each cross-attention layer, the text hidden state serves as the query and the visual features as keys and values, and the result is injected back into the backbone via a \emph{gated residual}:
\begin{equation}
  \mathbf{h} \leftarrow \mathbf{h} + \tanh(g_{\text{attn}})\cdot \mathrm{CrossAttn}(\mathbf{h},\,\mathbf{V}),\qquad
  \mathbf{h} \leftarrow \mathbf{h} + \tanh(g_{\text{ffn}})\cdot \mathrm{FFN}(\mathbf{h}),
  \label{eq:gated-residual}
\end{equation}
where $\mathbf{V}$ denotes the visual K/V and $g_{\text{attn}}, g_{\text{ffn}}$ are learnable gating scalars. The $\tanh$ gating makes the injection magnitude learnable and initially close to the identity map, so adding vision perturbs the base LLM only slightly and training is more stable.

\paragraph{Temporal positional encoding.} This is the key adaptation that makes cross-attention suitable for video time. Native cross-attention in Llama-3.2-Vision applies no rotary positional encoding to its queries or keys, leaving the visual side with no temporal position aligned to the text: even with the two-channel backbone in place, the visible frames remain mutually orderless on the visual side---the model has no way to tell which frame is earlier or later, which is a fundamental restriction for any video understanding that depends on time.

To address this, we equip \emph{both sides} of cross-attention with rotary positional encoding (RoPE)~\citep{su2024roformer}: the text query uses its text position, and each visual key uses the position of its \texttt{<|video|>} placeholder within the interleaved sequence of text and frames. In other words, text and frames share \emph{a single positional axis}: we number the entire interleaved sequence in one pass, text keeps its own indices, and each frame takes the index of its placeholder. Crucially, \emph{all visual tokens belonging to one frame share that same index}---cross-attention no longer distinguishes intra-frame spatial positions (intra-frame spatial structure is already carried by the vision encoder), and the positional signal collapses into the pure temporal signal of ``which frame this is''. This assignment relies on a \textbf{fixed resolution}: regardless of the original size, each frame is scaled to the same resolution before being fed to the vision encoder and produces the same number of visual tokens, so that ``one frame, one position'' is uniform across the full sequence and the temporal scale is regular. Frames and text then sit on a single temporal axis, the model can tell which frame comes first or last and align ``what happened in which frame'', and the positional basis required for temporal understanding is in place.

In addition, the \texttt{cross\_attention\_mask} controls the visibility of each \texttt{<|video|>} placeholder to its corresponding frame's visual features (the implementation of point~(5) above); when a new frame arrives or during step-by-step decoding, its visual K/V is appended to the per-layer cache (point~(2)), and already-encoded frame K/V is reused in subsequent steps (point~(3)).

\subsection{Vision encoding and compression}
\label{subsec:arch-encoder}

\paragraph{Vision encoder.} We use the ViT of Llama-3.2-Vision (patch size 14, image size 560, hidden 1280; 32 local layers + 8 global layers). Each frame yields about $40 \times 40$ patches (plus one CLS token); the encoder concatenates the outputs of 5 intermediate layers with the final global layer ($6 \times 1280 = 7680$ dimensions) as the per-frame visual feature.

\paragraph{Pooling compression (\texttt{get\_2dPool}).} After dropping the CLS token, the $H \times W$ patch grid is mean-pooled along its spatial dimensions with a given \texttt{stride}. The number of visual tokens per frame falls to roughly $1/\text{stride}^2$, which directly cuts the visual K/V injected into the LLM---saving memory and lightening every cross-attention step.

\paragraph{Projector.} A single linear layer projects the 7680-dimensional visual feature to the LLM hidden size $d_{\text{model}}{=}4096$ as the K/V input of cross-attention.

\section{Data, Training, and Inference}
\label{sec:data}

The architecture established earlier can look while it answers and answer while it looks, but the architecture by itself does not dictate \emph{when} the model should revise or stay silent. This behavior can only be acquired from data and training. A model trained solely on the ``whole clip, single answer'' corpora used for offline video understanding, even if its architecture supports continuous perception, will at inference time only produce a one-shot complete reply and then either repeat itself or stay silent at inappropriate moments---it has simply never seen any supervision in which the answer is rewritten as the stream advances.

\subsection{Data assets and training pipeline}
\label{subsec:data-overview}

\subsubsection*{Composition of the training data}

The training data fall into three groups, each addressing a different capability target.

\paragraph{Basic understanding data} teach the model the underlying vision--language alignment and form the bulk of the pre-training corpus, covering both English and Chinese. They split into four categories, each drawn from broad sources: \emph{image captions} (natural images, synthetic images, charts and documents); \emph{video captions} (web short clips, film and television, action and behavior, and egocentric footage); \emph{OCR} (both natural-scene text and synthetic documents, balanced across Chinese and English); and \emph{interleaved image-text}~\citep{zhu2023mmc4,laurencon2023obelics,wang2025unifiedvisual} (web-scale image-text, multimodal textbooks, and interleaved image-text reasoning data). These sources are not mixed indiscriminately---they are passed through a tiered quality filter, with low-quality sources discarded and original annotations replaced by recaptioned~\citep{chen2024sharegpt4v,chen2024sharegpt4video} versions where available, to lift the density and accuracy of the alignment corpus. In addition to these collected and recaptioned sources, we also synthesize our own high-quality \emph{hierarchical video captions}: for the same video we produce timestamped captions at three granularities, from coarse to fine---\emph{video}, \emph{event}, and \emph{action}---which respectively summarize the whole clip, describe event-level segments, and detail individual actions, jointly characterizing the content at multiple resolutions. These hierarchical captions also serve as the basis of the real-time data synthesis pipeline below.

\paragraph{Instruction data} give the model the ability to follow instructions, and form the bulk of the offline SFT corpus (about 8\,M samples) through large-scale collection of open-source data. Their coverage is broad: in modality, text-only, single-image, multi-image, and video, with both single- and multi-turn dialogues; in task, general QA, captioning, document / chart / interface understanding, OCR, math and multi-step reasoning, subject knowledge, code, and writing. The same tiered quality filtering applies.

\paragraph{Real-time synthesized data} teach the ``look-and-revise and timely-silence'' behavior, and are derived from the hierarchical captions above by the synthesis pipeline below; they are the supervision signal that distinguishes our work from standard video LLMs.

The first two follow well-established alignment and instruction-tuning recipes~\citep{liu2023llava,tan2025dpa} to first produce a general \emph{offline video understanding} model, and the third \emph{specializes} that offline model into a real-time one. This ``general first, specialize later'' division of labor determines the staging that follows.

\subsubsection*{The four training stages}

The full pipeline starts from \textbf{Llama-3.2-11B-Vision-Instruct} and advances through four stages (\Cref{tab:training-stages}), each continued from the previous checkpoint. All stages use next-token prediction, a cosine learning-rate schedule with a 0.1 warm-up ratio, DeepSpeed ZeRO-2~\citep{rajbhandari2020zero}, and one training epoch.

\begin{table}[t]
  \centering
  \caption{Goals, main data, scale, and trainable modules of the four training stages.}
  \label{tab:training-stages}
  \setlength{\tabcolsep}{7pt}
  \renewcommand{\arraystretch}{1.3}
  \begin{tabular}{@{}
      >{\raggedright\arraybackslash}p{2.5cm}
      >{\raggedright\arraybackslash}p{3.7cm}
      >{\raggedright\arraybackslash}p{2.8cm}
      >{\centering\arraybackslash}p{1.25cm}
      >{\raggedright\arraybackslash}p{3.7cm}@{}}
    \toprule
    \textbf{Stage} & \textbf{Goal} & \textbf{Main data} & \textbf{Scale} & \textbf{Trainable modules} \\
    \midrule
    \textbf{Stage~1}
      & Inject vision into the text channel; align the two modalities
      & Image--text pairs
      & 15\,M
      & Vision encoder, projector, and cross-attention layers\newline{\footnotesize\itshape (LLM backbone frozen)} \\
    \addlinespace
    \textbf{Stage~1.5}
      & Acquire video temporality on top of the alignment
      & 470\,K images, 1369\,K videos
      & 1.8\,M
      & Full model \\
    \addlinespace
    \textbf{Offline SFT}
      & General instruction following and offline video understanding
      & Text / image / video instructions
      & 8\,M
      & Full model \\
    \addlinespace
    \textbf{Real-Time SFT}
      & Look-and-revise behavior and the silence decision
      & Real-time synthetic $+$ offline QA mix
      & 836\,K
      & Full model \\
    \bottomrule
  \end{tabular}
\end{table}

Two staging choices warrant explanation.

\paragraph{Stage~1: train the bridge, freeze the backbone.} What needs to be learned at this stage is the bridge that turns vision into representations the LLM can consume---namely the vision encoder, the projector, and the cross-attention layers responsible for injection---while the self-attention backbone that carries language ability, together with the token embeddings, output head, and norms, remains frozen. This division of labor is designed to protect the language ability already present in the base LLM: with cross-attention initially close to the identity map ($\tanh$ gating) and the bridge not yet aligned, unfreezing the text backbone at the same time would let misaligned visual gradients perturb or even degrade the language ability. Once alignment is established (from Stage~1.5 onward) we unfreeze everything so that the entire model can co-adapt to video and to longer contexts.

\paragraph{Offline SFT first, then a real-time SFT.} The real-time synthetic data (836\,K) is small in scale and narrow in distribution compared with the offline instruction data (8\,M). Training directly on the smaller corpus from scratch risks hurting the general understanding that the offline stage has built up. Real-time SFT therefore continues from the offline SFT checkpoint, the learning rate is lowered from $1\mathrm{e}{-5}$ to $5\mathrm{e}{-6}$, and we use a larger gradient-accumulation step (4) for more stable updates---positioning it as a \emph{specialization} that injects real-time behavior on top of a robust offline model rather than as a from-scratch training run.

\subsection{Real-time data synthesis}
\label{subsec:data-synth}

\paragraph{Motivation.} Real-time behavior has three traits: continuous perception, prompt revision the moment evidence appears, and silence when there is nothing to say. All three require \emph{supervision}, and off-the-shelf data does not provide it:
\begin{itemize}[leftmargin=1.4em, itemsep=2pt, topsep=2pt]
  \item Static captions and QA are ``whole clip $\rightarrow$ single answer'' and contain no trajectory in which the answer is rewritten over time.
  \item They also lack \texttt{<|silence|>}---the model has no way to learn ``stay silent now''.
\end{itemize}
The samples needed for real-time training have a distinctive shape: \emph{the instruction is fixed, but its best response evolves as the video stream advances}. For instance, the instruction ``What is the score now?'' itself does not change, yet as a goal is scored the correct reply must be rewritten; between two goals, the model should remain silent. Such data require the model to ``revise as it watches, and answer only when evidence allows''---the very target of real-time training, which static QA can neither evaluate nor train for.

\paragraph{Basis for synthesis: hierarchical captions.} The starting point of our synthesis is not the raw video but the hierarchical captions introduced earlier. This multi-granularity dense description transcribes a video into a structured timeline of text---fine enough to expose useful signals for change-point detection, and precise enough that the instructions constructed next can be anchored to specific moments. The real-time QA is synthesized on top of this representation.

\begin{figure}[t]
  \centering
  \includegraphics[width=\linewidth]{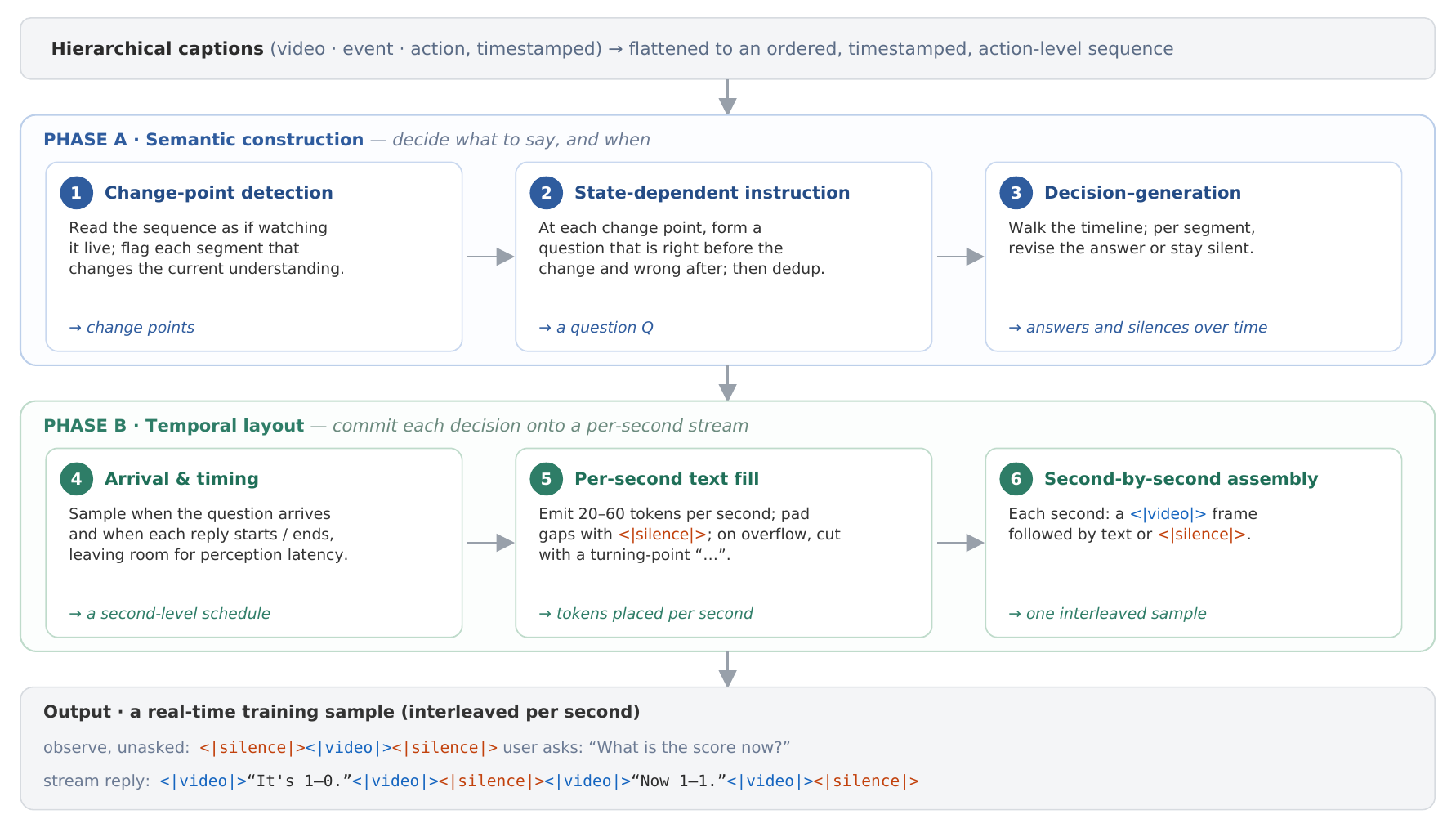}
  \caption{The real-time data synthesis pipeline. Starting from the hierarchical captions (flattened to an ordered, timestamped, action-level sequence), \textbf{Phase~A (semantic construction)} decides \emph{what to say and when}---change-point detection, state-dependent instruction generation, and a decision--generation response that either revises the answer or stays silent. \textbf{Phase~B (temporal layout)} commits these decisions onto a per-second stream---sampling the question's arrival and each reply's start/end, filling $20$--$60$ tokens per second, and assembling a \texttt{<|video|>}--text/\texttt{<|silence|>} sequence. The product is a real-time training sample whose answer is rewritten as the stream advances.}
  \label{fig:synthesis}
\end{figure}

The synthesis proceeds in two phases (\Cref{fig:synthesis}). \textbf{Semantic construction} decides ``when, about what, to speak or to stay silent''; \textbf{temporal layout} then commits each of these decisions onto a per-second multimodal sequence, which is finally assembled into multi-turn samples.

\subsubsection*{Semantic construction: from hierarchical captions to a stream-evolving response sequence}

We flatten the hierarchical captions to action granularity and obtain a temporally ordered description sequence $\{(t_i, c_i)\}$, where $c_i$ is the text description of the $i$-th action / scene segment and $t_i = [s_i, e_i]$ is its time interval. Semantic construction operates on this sequence in three steps.

\paragraph{Key change-point detection.} We have a large language model examine the sequence as if it were watching it in real time: for each $c_n$, the model compares it with all earlier $c_{<n}$ and decides whether it introduces information sufficient to alter the current understanding, classifying it into one of three categories---turning point (state reversal), disambiguation / completion, or external context change (the first description has no prior context and is never counted as a change). The set of detected change points $\{k\}$ characterizes ``the moments at which the world has changed enough to merit a revision''---the anchors for the subsequent steps.

\paragraph{State-dependent instruction generation.} At each change point $k$, with the context before the change $c_{<k}$ taken as the known information and $\Delta_k$ the new information, the model is prompted to generate a user instruction $Q$. Unlike generic visual QA, these instructions are constrained to be \emph{state-dependent}: $Q$ has a clearly correct answer before $\Delta_k$ occurs, and that answer becomes incorrect or incomplete immediately after $\Delta_k$; in addition, $Q$ must focus on attributes of the scene that truly vary over time (actions, relative positions, interactions), not on static facts that hold throughout (color, material, intrinsic identity). To preserve diversity, the surface form of the instruction is sampled along three axes---syntactic structure, expressive style, and emotional tone---according to preset distributions, and the instructions generated across change points are then aggregated and deduplicated by cosine similarity in a semantic embedding space.

\paragraph{Decision--generation response (iterative update and silence).} Given an instruction $Q$, we first synthesize an anchor response $a_1$ based on the information up to the change point, as the answer to $Q$ at the initial moment. The model then walks the timeline segment by segment from the change point onward: for each subsequent segment it makes a binary decision---does this segment alter the current best response to $Q$? If \emph{yes}, it synthesizes an updated response $a_{k+1}$ from all information up to that point; if \emph{no}, it emits \texttt{Silence}, indicating that the model should stay silent here. A key design point: \emph{silent segments are not written into the response history}, so the model is not biased toward silence on later decisions by its own prior silence; a task that has already been answered is also explicitly marked, raising the bar for whether the model should speak again after the answer is given. Iterating to the end of the sequence yields the pair $(Q;\; a_1, a_2, \dots)$, where each $a_k$ is either a real text answer or \texttt{Silence}. This single step teaches the model both when to revise (a new $a_{k+1}$) and when to stay silent.

\subsubsection*{Temporal layout: aligning responses to a frame-by-frame stream}

The pairs $(Q;\; a_1, a_2, \dots)$ produced by the semantic phase are still segment-level---they cannot be used directly for training. The model, at inference time, reads frames continuously at 1\,fps and emits tokens one by one, so its supervision sequence must commit to ``what should be emitted right after each frame''. The temporal-layout phase lays out each response onto a timeline with a 1-second step, deciding second by second whether to follow each \texttt{<|video|>} placeholder with text or with \texttt{<|silence|>}. Let the triggering segment of response $a_k$ be $[s_k, e_k]$ (in seconds).

\paragraph{Instruction arrival time.} Let the triggering segment of the first response $a_1$ (i.e., the segment immediately before the key change) be $[s_1, e_1]$. The arrival time $t_q$ of the user instruction $Q$ is sampled uniformly within the first 80\% of this segment,
\begin{equation}
  t_q \sim \mathcal{U}\!\left(s_1,\ s_1 + 0.8\,(e_1 - s_1)\right).
\end{equation}
This guarantees that $Q$ is asked before the change has occurred, so that the initial answer is later overturned by new information---exactly the shape this construction is meant to capture.

\paragraph{Start time of each response.} The start time $p_k$ of each non-silent response $a_k$ is sampled uniformly within the \emph{last third} of its triggering segment,
\begin{equation}
  p_k \sim \mathcal{U}\!\left(e_k - \tfrac{1}{3}(e_k - s_k),\ e_k\right),
\end{equation}
with the additional constraint $p_1 \ge t_q$ for the first response. That is, the model does not start replying at the segment boundary but only after it has watched the segment and confirmed the change, implicitly encoding the latency required for perception and decision; the gap $[t_q, p_1)$ between the question arriving and the first response starting is filled with silence accordingly.

\paragraph{End time of each response.} The end time $q_k$ is determined by ``the next utterance'', so neighboring responses meet head-to-tail in time. If $a_k$ is immediately followed by another non-silent response $a_{k'}$, then $q_k = p_{k'}$. If it is followed by a silent segment $[s_m, e_m]$, then $a_k$ is allowed to extend across the segment boundary into the first half of the silent segment before stopping, with $q_k \sim \mathcal{U}\!\left(s_m,\ s_m + 0.5\,(e_m - s_m)\right)$. The last response extends to the end of the video. Whenever the start times of neighboring responses would conflict, they are forcibly staggered by at least one second.

\paragraph{Text fill between start and end.} Given the interval $[p_k, q_k)$, we first tokenize $a_k$ and then fill it second by second: each second consumes a number of tokens sampled uniformly from $[20, 60]$ (rounded at subword boundaries to keep them decodable), simulating a token-by-token streaming pace; the random per-second budget also makes the data robust to a \emph{variable} decoding throughput at inference time, rather than overfitting to a single pace. Two boundary cases follow naturally. First, if the text has finished within the budget, the remaining seconds are filled with \texttt{<|silence|>} and an additional \texttt{<|silence|>} is appended at the end of the response to mark the end of this utterance. Second, if the text has not finished within the interval, the response is \emph{truncated}, the overflow is recorded separately, and a \textbf{turning-point token} is inserted at the start of the subsequent segment. The latter corresponds to a key phenomenon of real-time interaction: the reply has not yet finished but the world has already changed, and the model uses the turning-point token (decoded as ellipsis ``\dots'') to interrupt the current utterance and pivot to the new response.

\paragraph{Second-by-second assembly.} The timeline advances from $t_q$ to the end of the video $T$: at every second, we first place a \texttt{<|video|>} placeholder, then determine which active response interval $[p_k, q_k)$ this second falls in and fill in the corresponding text fragment; if this second is not covered by any response (the gap before the first reply, a silence-decision segment, or the tail after a reply has finished), we fill in \texttt{<|silence|>}. The result is a per-second interleaved sequence of the form \texttt{<|video|>}--text/\texttt{<|silence|>}--\texttt{<|video|>}--text/\texttt{<|silence|>}--\dots, encoding the full real-time trajectory of ``keep perceiving, speak when appropriate, correct when necessary, and stay silent the rest of the time'' into a single sample.

\subsubsection*{Multi-turn assembly and output format}

Finally, multiple real-time QA constructed on the same video are concatenated into a multi-turn dialogue with a system prompt (the system prompt fixes the real-time persona; the full text appears below). Each dialogue begins with a pure-silence prelude from the start of the video to the instruction arrival time $t_q$---an interleaving of \texttt{<|silence|>} and \texttt{<|video|>} that signals ``the user has not yet asked, but the model is already observing''---followed by alternations of user instructions and the model's real-time streaming reply. \emph{Two-turn} samples carry a single QA; \emph{three-turn} samples concatenate two adjacent QAs and, at the moment the second instruction arrives, truncate the reply to the first QA at a randomly chosen ratio, depicting the realistic situation in which ``the previous answer has not yet finished but a new instruction has arrived''. The mixing ratio of two- to three-turn samples is set to a target value, and we make sure no QA is discarded.

Each sample carries a video field strictly aligned to the number of frames, recording the source video and the timestamp of every frame, one for each \texttt{<|video|>} in the sequence. When the total duration of a sample exceeds the frame budget (determined by sampling rate, maximum, and minimum frame count), excess frames are trimmed from the tail; samples that are still too long are skipped entirely. The final product is real-time interaction supervision ready for real-time SFT, with a format that maps one-to-one onto the formalization given earlier.

\subsection{Real-time SFT and the silence decision}
\label{subsec:rt-sft}

Real-time SFT is continued from the offline SFT model and makes it acquire the real-time behavior synthesized earlier: training mixes real-time and offline QA, system prompts distinguish the two modes, and at inference time a single threshold gates when to stay silent.

\subsubsection*{Training data and mode separation}

Real-time SFT is not trained on the real-time synthetic data alone but on a \emph{mixture} of that data with offline QA. Mixing in offline data is meant to preserve general understanding while specializing for real-time behavior: training on real-time data alone leads the model to overfit ``always observing, frequently silent'' as its only mode, at the cost of general offline QA ability.

The real-time and offline modes are two \emph{starkly different} working regimes, and we distinguish them by \emph{different system prompts} so that the same model toggles its behavior according to the prompt. The real-time mode requires continuous perception, observation-grounded answering, prompt revision the moment a relevant change occurs, and \texttt{<|silence|>} when there is nothing to say:

\begin{promptbox}{Real-time system prompt}
You are a helpful AI assistant. You perceive and understand the surrounding environment in real-time through the camera and interact with the user. Whether or not the user actively asks questions, you continuously observe and analyze visual information, maintaining awareness of the environment.

\smallskip
\noindent\textbf{Core Abilities:}
\begin{itemize}[leftmargin=1.4em, itemsep=1pt, topsep=2pt]
  \item \textbf{Continuous Perception:} Always observe and analyze the visual information captured by the camera in real-time. This perception does not stop even when there is no user interaction.
  \item \textbf{Observation-Based Answers:} All answers must be based on visual observations, including both historical and real-time data. Do not make guesses without evidence from visual input.
  \item \textbf{Dynamic Adjustment:} When you observe changes relevant to the user's question, promptly update and adjust your answers to ensure timeliness and accuracy.
\end{itemize}

\smallskip
\noindent\textbf{Interaction Rules:}
\begin{itemize}[leftmargin=1.4em, itemsep=1pt, topsep=2pt]
  \item Always base your answers on observed visual information.
  \item If you notice significant changes related to the user's question, proactively and promptly update your answer.
  \item When you are unable to answer or have finished answering, output \texttt{<|silence|>} directly.
\end{itemize}

\smallskip
Your goal is to be the user's reliable ``eyes'', helping them understand and perceive the world around them.
\end{promptbox}

The offline mode is the conventional QA form---answer according to the given text, image, or video, with neither continuous perception nor silence:

\begin{promptbox}{Offline system prompt}
You are a helpful AI assistant. Respond to the user's request based on the provided text, images and/or videos.
\end{promptbox}

This ``one model, two prompts, two modes'' design also provides the data foundation for the two inference entry points below.

\subsubsection*{Silence decision: inference-time threshold gating}

In the real-time data, \texttt{<|silence|>} is the dominant token across frames, and the model has to decide ``answer now or stay silent now'' at almost every frame. We delegate this decision to inference time and control it with a single rule: at each step we take the predicted probability of \texttt{<|silence|>}, and we allow the model to be silent only when this probability is at least a threshold (the released code uses 0.6); otherwise the silence probability is zeroed out and the distribution is re-normalized, so that the model is forced to speak. This bias makes the model lean toward timely responses rather than excessive silence (the full gating mechanism, together with inference-time handling of the turning-point token, is presented below).

We find that as long as the real-time data synthesis is of sufficient quality, no additional training-time loss weighting on \texttt{<|silence|>} is needed---this inference-time threshold alone is enough to give the model strong real-time behavior.

\subsection{Real-time inference: an event-driven online loop}
\label{subsec:rt-infer}

What real-time SFT trains is a \emph{behavior}: decide per frame whether to answer or to stay silent, and revise the answer the moment a relevant change appears. At inference time this behavior is realized as an \emph{event-driven online loop}: frames keep arriving at 1\,fps, the user may ask at any moment, and the model toggles between two states---\textsc{Waiting} and \textsc{Replying}---with \texttt{<|silence|>} regulating the transitions. Frame reading and token generation advance along two concurrent pathways without blocking each other (the system-level counterpart of the two-channel design), so that a frame arriving mid-reply only needs its K/V appended to the cache to be available to the subsequent generation; no decoding step has to be interrupted in order to ingest it. New frames sit on the same unified position axis as in training: each arriving frame appends its \texttt{<|video|>} placeholder to the sequence and takes the index of that placeholder (rather than starting a separate counter), so the temporal axis stays monotonically continuous throughout the stream and the temporal encoding is consistent between training and inference.

\paragraph{The two-state loop (\Cref{alg:rt-loop}).} The loop is driven by \emph{input events}---\emph{a new frame arrives or a new user question is posed; either suffices on its own}. Each such event injects the input into the context (a frame's visual features are appended to the K/V caches of all cross-attention layers; a question is appended to the text sequence), and the model immediately predicts the next token, adjudicating ``answer now, or emit \texttt{<|silence|>}'' (silence is allowed only when its probability is at least the threshold $\tau{=}0.6$). The two states then arise naturally. In \textsc{Waiting}, the model has either finished answering or has nothing to report; for each new frame or question it runs the above prediction, stays in \textsc{Waiting} without decoding if the verdict is \texttt{<|silence|>}, and transitions to \textsc{Replying} otherwise. In \textsc{Replying}, the model autoregressively emits reply tokens until it outputs \texttt{<|silence|>}, which sends it back to \textsc{Waiting}. We emphasize that \emph{events arriving mid-reply trigger the same prediction}: new frames' K/V is continually appended, new questions are inserted immediately, and the model either \emph{continues} the current reply, \emph{interrupts} the previous answer with a turning-point token when the previous answer is overturned and pivots to the new evidence, or transitions directly to silence.

\begin{algorithm}[t]
  \caption{Real-time inference loop (event-driven, silence-gated).}
  \label{alg:rt-loop}
  % Pin comments to a fixed-width right-hand column so every ``>'' starts at the
  % same x; the parbox lets long comments wrap tidily inside that column.
  \algrenewcommand{\algorithmiccomment}[1]{\hfill\(\triangleright\)~\parbox[t]{4.6cm}{\raggedright #1}}
  \begin{algorithmic}[1]
    \Require silence threshold $\tau = 0.6$
    \State $\texttt{state} \gets \textsc{Waiting}$
    \Loop
      \If{$\texttt{state} = \textsc{Waiting}$} \Comment{block until a frame or a question arrives}
        \If{a new frame arrives}
          \State encode it; append its K/V to the per-layer cache \Comment{previous frames' K/V is reused, not recomputed}
        \EndIf
        \If{a new question arrives} \Comment{independent of the frame; if both arrived, inject both}
          \State append it to the text sequence
        \EndIf
        \State forward to obtain the next-token distribution $p$
        \If{$p(\texttt{<|silence|>}) \ge \tau$}
          \State stay \textsc{Waiting} \Comment{nothing to say; keep observing}
        \Else
          \State $\texttt{state} \gets \textsc{Replying}$
        \EndIf
      \Else \Comment{\textsc{Replying}: emit reply tokens as events keep arriving}
        \If{a new frame arrives}
          \State encode it; append its K/V to the cache \Comment{previous frames reused}
        \EndIf
        \If{a new question arrives} \Comment{independent; a frame and a question may co-arrive}
          \State insert it into the text sequence
        \EndIf
        \State emit the next token---may continue the reply, switch via a turning-point token, or end the reply
        \If{the emitted token is \texttt{<|silence|>}}
          \State $\texttt{state} \gets \textsc{Waiting}$
        \EndIf
      \EndIf
    \EndLoop
  \end{algorithmic}
\end{algorithm}

\paragraph{Two inference entry points.} A single set of weights exposes two generation entry points sharing the K/V caching mechanism above, differing only in \emph{whether frames arrive continuously or are supplied all at once}: \texttt{real\_time\_generate} executes the event-driven loop above, and \texttt{offline\_generate} takes the full video and question at once and produces the entire reply, which we use for offline evaluation and as a control.\footnote{The inference code (\texttt{offline\_generate} / \texttt{real\_time\_generate} and the example inference scripts) is released alongside the model weights at the HuggingFace collection \href{https://huggingface.co/collections/OpenMOSS-Team/moss-video-preview}{OpenMOSS-Team/moss-video-preview}; the streaming real-time variant corresponds to \texttt{moss-video-preview-realtime-sft}.} The quantitative inference latency and throughput are reported later.

\section{Experiments}
\label{sec:exp}

This section answers experimentally the three questions this work, as a preview, must address:
\begin{enumerate}[label=(\arabic*), leftmargin=1.6em, itemsep=2pt, topsep=2pt]
  \item \textbf{Usability.} Does a model trained on a cross-attention backbone---taken as a conventional (offline) video and multimodal understanding model---have competitive capability?
  \item \textbf{Efficiency.} Does the architectural prediction---``lower frequency of visual processing $\Rightarrow$ faster inference''---hold up in measurement?
  \item \textbf{Cost.} Does specializing the model from offline to real-time (the real-time SFT, which introduces silence and dynamic revision) come at the price of offline understanding ability?
\end{enumerate}

\emph{Real-time interaction itself has no standardized quantitative benchmark yet}, for the reasons set out earlier. The quantitative results in this section therefore focus on ``offline understanding'' and ``inference efficiency'', and the real-time capability is presented qualitatively through demonstrations.

\subsection{Experimental setup}
\label{subsec:setup}

\paragraph{Models under evaluation.} We report two checkpoints of our model: \textbf{offline SFT} (the version before real-time specialization) and \textbf{real-time SFT} (the released real-time version). Reporting both is meant to separate ``general understanding ability'' from ``the cost of real-time specialization''.

\paragraph{Comparison models.} We use three points of comparison:
\begin{itemize}[leftmargin=1.4em, itemsep=2pt, topsep=2pt]
  \item \textbf{Llama-3.2-11B-Vision}~\citep{grattafiori2024llama3}, the \emph{starting base} of this work, used to gauge the gain (and the change) brought by our training relative to the base;
  \item \textbf{LLaVA-OneVision-1.5-8B-Instruct}~\citep{an2025llavaonevision15}, a representative decoder-only open-source multimodal model;
  \item \textbf{Qwen2.5-VL-7B-Instruct}~\citep{bai2025qwen25vl}, a strong open-source reference at comparable (and in fact smaller) scale, taken as the principal baseline for this section---``gap'' and ``ahead'' below are measured against it.
\end{itemize}

\paragraph{Benchmarks.} We cover 24 public benchmarks in four capability categories: \emph{Doc / OCR} (OCRBench~\citep{liu2024ocrbench}); \emph{Multimodal perception} (MMStar~\citep{chen2024mmstar}, MMBench-CN/EN~\citep{liu2024mmbench}, MMMU~\citep{yue2024mmmu}, RealWorldQA~\citep{xai2024realworldqa}, MuirBench~\citep{wang2024muirbench}, SEEDBench~\citep{li2023seedbench}, MME-RealWorld~\citep{zhang2025mmerealworld}, POPE~\citep{li2023pope}, CV-Bench~\citep{tong2024cambrian}, V$^*$~\citep{wu2024vstar}); \emph{Multimodal reasoning} (AI2D~\citep{kembhavi2016ai2d}, VisuLogic~\citep{xu2025visulogic}, VLMsAreBlind~\citep{rahmanzadehgervi2024vlmsblind}, ZeroBench~\citep{roberts2025zerobench}); \emph{Video understanding} (VideoMME~\citep{fu2025videomme}, EgoSchema~\citep{mangalam2023egoschema}, LongVideoBench~\citep{wu2024longvideobench}, MLVU~\citep{zhou2025mlvu}, LVBench~\citep{wang2025lvbench}, TempCompass~\citep{liu2024tempcompass}, VSI-Bench~\citep{yang2025vsibench}, Video-Holmes~\citep{cheng2025videoholmes}).

\paragraph{Efficiency setup.} Inference efficiency is measured on a \textbf{single NVIDIA H200} with $256$ frames sampled from the same video; both models use bf16 + FlashAttention-2~\citep{dao2023flashattention2} and greedy decoding (\texttt{do\_sample=False}). The reported metrics are \textbf{TTFT} (time to first token, prefill inclusive), \textbf{TPS} (decoding throughput, the steady-state per-token rate after prefill), \textbf{end-to-end total latency}, and \textbf{P95 TTFT}; we run one warm-up and then take the mean (and P95) over multiple runs. We emphasize that this is a \emph{single-video, single-configuration} speed comparison rather than a standardized benchmark suite.

\subsection{General video and multimodal understanding}
\label{subsec:general}

\Cref{tab:bench} gives per-benchmark scores in the four categories. The first two columns are our two checkpoints (in \textbf{bold}); ``--'' marks the benchmarks on which a model has not been reported.

\begin{table}[t]
  \centering
  \small
  \setlength{\tabcolsep}{6.5pt}
  \renewcommand{\arraystretch}{1.25}
  \caption{General multimodal and video understanding evaluation (higher is better; OCRBench is on a 0--1000 scale; all others are percentage scores). \textbf{Bold} marks the two checkpoints of our model.}
  \label{tab:bench}
  \begin{tabular}{@{}llcccccc@{}}
    \toprule
    \textbf{Category} & \textbf{Benchmark}
    & \makecell{\textbf{real-time}\\\textbf{SFT}}
    & \makecell{\textbf{offline}\\\textbf{SFT}}
    & \makecell{Llama-3.2\\11B-Vision\\(base)}
    & \makecell{LLaVA-OV\\1.5-8B}
    & \makecell{Qwen2.5-VL\\7B} \\
    \midrule
    Doc / OCR
      & OCRBench           & \textbf{677.00} & \textbf{705.00} & 759.00 & 829.00 & 864.00 \\
    \midrule
    \multirow{11}{*}{\makecell[l]{Multimodal\\perception}}
      & MMStar             & \textbf{53.11}  & \textbf{48.99}  & 53.87  & 67.72  & 63.90  \\
      & MMBench-CN (dev)   & \textbf{83.04}  & \textbf{82.03}  & 68.03  & 81.00  & 86.07  \\
      & MMBench-EN (dev)   & \textbf{83.97}  & \textbf{83.09}  & 72.76  & 84.14  & 87.76  \\
      & MMMU               & \textbf{47.36}  & \textbf{45.26}  & 41.70  & 55.44  & 54.90  \\
      & RealWorldQA        & \textbf{60.92}  & \textbf{59.48}  & 66.27  & 68.10  & 69.28  \\
      & MuirBench          & \textbf{39.92}  & \textbf{39.88}  & --     & 37.50  & 45.42  \\
      & SEEDBench          & \textbf{69.40}  & \textbf{50.10}  & --     & 77.32  & 74.00  \\
      & MME-RealWorld      & \textbf{44.30}  & \textbf{51.89}  & --     & 62.31  & 54.65  \\
      & POPE               & \textbf{88.17}  & \textbf{87.88}  & --     & 89.20  & 87.68  \\
      & CV-Bench           & \textbf{73.05}  & \textbf{70.17}  & --     & 80.82  & 80.14  \\
      & V$^*$              & \textbf{49.74}  & \textbf{62.83}  & --     & 73.30  & 71.73  \\
    \midrule
    \multirow{4}{*}{\makecell[l]{Multimodal\\reasoning}}
      & AI2D               & \textbf{77.33}  & \textbf{75.06}  & 76.46  & 84.16  & 83.03  \\
      & VisuLogic          & \textbf{28.60}  & \textbf{28.70}  & --     & 27.00  & 25.90  \\
      & VLMsAreBlind       & \textbf{50.21}  & \textbf{47.48}  & --     & 51.07  & 52.32  \\
      & ZeroBench (Sub)    & \textbf{7.83}   & \textbf{8.53}   & --     & 11.98  & 8.99   \\
    \midrule
    \multirow{9}{*}{\makecell[l]{Video\\understanding}}
      & VideoMME              & \textbf{62.48} & \textbf{59.81} & --   & --   & 65.10  \\
      & EgoSchema (subset)    & \textbf{54.80} & \textbf{47.40} & --   & --   & 63.80  \\
      & LongVideoBench        & \textbf{51.61} & \textbf{54.08} & --   & --   & 54.70  \\
      & MLVU (dev)            & \textbf{61.81} & \textbf{60.32} & --   & --   & 70.20  \\
      & LVBench               & \textbf{38.93} & \textbf{39.70} & --   & --   & 45.30  \\
      & TempCompass (MC)      & \textbf{59.68} & \textbf{61.65} & --   & --   & 72.53  \\
      & TempCompass (Y/N)     & \textbf{72.03} & \textbf{70.73} & --   & --   & 74.36  \\
      & VSI-Bench             & \textbf{36.20} & \textbf{33.48} & --   & --   & 28.30  \\
      & Video-Holmes          & \textbf{39.30} & \textbf{39.50} & --   & --   & 33.00  \\
    \bottomrule
  \end{tabular}
\end{table}

\paragraph{Relative to the base: training brings instruction-following and multimodal QA ability.} On the items reported by the base Llama-3.2-11B-Vision, our improvements concentrate on multimodal instruction-style QA: MMBench-EN rises from 72.76 to 83.97 (real-time, $+11.2$), MMBench-CN from 68.03 to 83.04 ($+15.0$), and MMMU from 41.70 to 47.36 ($+5.7$). This shows that the cross-attention training pipeline and instruction tuning have indeed turned a native Llama-3.2-Vision base into a multimodal model capable of following instructions and answering. The cost is also clearly visible: OCRBench drops (759 $\rightarrow$ 677/705) and RealWorldQA drops (66.27 $\rightarrow$ 60.92/59.48)---in shifting the capability emphasis toward instruction following and video temporality, training has traded away part of the pure-OCR and fine-grained real-world perception ability.

\paragraph{Relative to the strong baseline Qwen2.5-VL-7B: a real gap.} On most perception, OCR, and video QA benchmarks, our model trails Qwen2.5-VL-7B, which is at a comparable (and in fact smaller) scale: OCRBench (677 vs.\ 864), EgoSchema (54.80 vs.\ 63.80, $-9.0$), MLVU (61.81 vs.\ 70.20, $-8.4$), TempCompass-MC (59.68 vs.\ 72.53, $-12.9$), and so on. For a preview, this trade-off reflects a stated position: first establish the paradigm and the architecture; general benchmark numbers are left to subsequent scaling of data and parameters.

\paragraph{The dimensions on which we lead align with the real-time main line.} On the items that require reasoning rather than memorization, our model overtakes Qwen2.5-VL-7B: \textbf{VisuLogic} on visual logical reasoning (28.60/28.70 vs.\ 25.90), \textbf{VSI-Bench} on visual-spatial intelligence (36.20 vs.\ 28.30, $+7.9$), and \textbf{Video-Holmes} on fine-grained spatio-temporal reasoning (39.30/39.50 vs.\ 33.00, $+6.3$). The common theme of these three is ``understanding what is happening and inferring from it''---spatial relations, action logic, and causality unfolding over time---which is exactly the capability dimension that matters most for real-time video understanding. In other words, our model falls short on the memorization-style general benchmarks yet leads on the reasoning-style benchmarks that align with its design objective; the resulting pattern reflects choices in architecture and data orientation rather than chance.

\paragraph{Key finding: real-time specialization comes at almost no cost to offline understanding.} Comparing our two columns (real-time SFT vs.\ offline SFT): the per-benchmark trends agree closely, and on several video benchmarks the real-time version is actually \emph{higher} (VideoMME 62.48 vs.\ 59.81; EgoSchema 54.80 vs.\ 47.40; VSI-Bench 36.20 vs.\ 33.48). On a few individual benchmarks the two versions diverge meaningfully (SEEDBench is about 19 points higher on real-time, V$^*$ is about 13 points higher on offline), but the overall mean and per-item trend track each other very closely. This answers the third question of this section: the silence decision and dynamic revision do not come at the cost of offline understanding---a point central to the argument of this work, because it shows that ``real-time behavior'' can be added as a nearly lossless specialization on top of a strong offline model rather than as a zero-sum trade-off.

\subsection{Inference efficiency}
\label{subsec:efficiency}

We argued that cross-attention lowers the frequency of visual processing and should yield faster inference. \Cref{tab:speed} gives the measured result for that prediction.

\begin{table}[t]
  \centering
  \setlength{\tabcolsep}{6pt}
  \renewcommand{\arraystretch}{1.2}
  \caption{Inference speed (single H200, 256 frames, same video and decoding configuration; higher TPS and lower latency are better).}
  \label{tab:speed}
  \begin{tabular}{@{}lcccccc@{}}
    \toprule
    \textbf{Model} & \textbf{Frames} & \textbf{Params}
    & \makecell{\textbf{Avg TTFT}\\(s)\,$\downarrow$}
    & \makecell{\textbf{Avg TPS}\\(tok/s)\,$\uparrow$}
    & \makecell{\textbf{Avg total}\\\textbf{latency} (s)\,$\downarrow$}
    & \makecell{\textbf{P95 TTFT}\\(s)\,$\downarrow$} \\
    \midrule
    \rowcolor{oursgray}\textbf{MOSS-Video-Preview} & 256 & 11B
      & \bnum{1.95} & \bnum{38.41} & \bnum{28.51} & \bnum{1.96} \\
    Qwen2.5-VL-7B   & 256 & 7B
      & 9.94 & 14.26 & 52.76 & 9.96 \\
    \bottomrule
  \end{tabular}
\end{table}

Under this configuration, our model achieves \textbf{about a $5\times$ TTFT speedup} (1.95\,s vs.\ 9.94\,s) relative to Qwen2.5-VL-7B, \textbf{about a $2.7\times$ decoding throughput} (38.41 vs.\ 14.26 tokens/s), and roughly a \textbf{46\% reduction} in end-to-end total latency (28.5\,s vs.\ 52.8\,s). Notably, our model has \textbf{11B} parameters and Qwen2.5-VL has \textbf{7B}---we are faster on every metric \emph{despite being the larger model}.

\paragraph{Why it is faster.} This is not the result of any engineering trick but follows from the architecture. Decoder-only models pour the large set of visual tokens from $256$ frames \emph{into the autoregressive sequence}: on the one hand, prefill must build the full self-attention K/V for this ultra-long sequence, so TTFT is high ($\approx 10$\,s); on the other, every subsequent decoding step then has to attend to a context containing all those visual tokens, so per-token decoding stays slow (TPS is low). Cross-attention, by contrast, \emph{lifts vision out of} the autoregressive sequence: the visual K/V is encoded once during prefill and is retrieved only in 8 of 40 layers, and decoding steps no longer carry a large pool of visual tokens through self-attention---prefill is lighter (lower TTFT) and per-step decoding is cheaper (higher TPS). Both advantages share the same source---``vision does not enter the autoregressive sequence''---which is the structural difference of the two designs.

\paragraph{Measurement boundary.} This is a single-video, single-configuration comparison, not a standardized throughput ranking across models; it is intended to corroborate the architectural argument above. The speeds are measured along the standard HuggingFace inference path (bf16 + FlashAttention-2), without any custom inference engine or serving acceleration, so the advantage can be attributed directly to the architecture rather than to an engineered serving stack.

\subsection{Real-time capability: qualitative}
\label{subsec:rt-qualitative}

As discussed earlier, real-time interaction still lacks a benchmark that \emph{jointly} measures accuracy and timeliness. Until one exists, we present the real-time capability \emph{qualitatively}, without a quantitative score.

We publish three demonstrations, one per inference entry point: \textbf{streaming real-time} (frames are fed in continuously and the model looks and answers concurrently, demonstrating the behavior described earlier---\texttt{<|silence|>} during quiet observation, an answer or revision when a relevant change occurs, and a turning-point token to interrupt a stale answer when needed); \textbf{offline video} and \textbf{offline image} (the full input is given at once and the entire reply is generated, corresponding to the offline entry point).\footnote{The three demonstrations (streaming real-time / offline video / offline image) are released alongside the model; see the ``Demo'' section of the GitHub release \href{https://github.com/OpenMOSS/MOSS-Video-Preview}{OpenMOSS/MOSS-Video-Preview}.} Among them, the streaming real-time demo most directly embodies what distinguishes our model from offline / streaming counterparts: it turns the ``keep perceiving during reply, revise on the spot'' behavior into observable interaction rather than leaving it at the level of a sequence format.

A qualitative demonstration can show the behavior succeeding but \emph{cannot quantify} how timely the revisions are or how appropriate the silences are---turning these judgments into reproducible metrics is an open problem we leave to future work.

\subsection{Analysis and discussion}
\label{subsec:discussion}

Taking these results together, the three questions of this section each have an answer, and together they support the position-paper stance of this work:
\begin{itemize}[leftmargin=1.4em, itemsep=2pt, topsep=2pt]
  \item \textbf{Usability (\cmark).} The cross-attention backbone and the training pipeline can take a native Llama-3.2-Vision base and produce \emph{a competitive offline video and multimodal understanding model}---a significant gain on instruction-style QA relative to the base, and overall capability comparable to a strong open-source model at a similar scale.
  \item \textbf{Efficiency (\cmark, and structurally so).} The roughly $5\times$ TTFT and $2.7\times$ TPS advantages persist with a larger parameter count, and are explainable directly by ``vision does not enter the autoregressive sequence''---the core motivation for the cross-attention choice is borne out in measurement.
  \item \textbf{Cost (\cmark, near-zero).} Real-time specialization does not measurably sacrifice offline understanding; on some video benchmarks the real-time version is even better---suggesting that ``real-time'' can be added as a specialization on top of a strong offline model.
\end{itemize}

The performance pattern of our model splits clearly along the type of benchmark: on knowledge-intensive and fine-grained perception benchmarks (OCR, RealWorldQA, parts of video QA) it trails Qwen2.5-VL, whereas on reasoning benchmarks (VisuLogic, VSI-Bench, Video-Holmes) it leads. The former depend mainly on the scale and quality of general pre-training data and a larger parameter scale---directions this preview has not yet invested in---while the latter depend on understanding the spatial and temporal structure of the scene, which aligns with the central demand of real-time video understanding. Overall, the gap on general benchmarks should be attributed to \emph{data and scale, not to the cross-attention architecture itself}.

\section{Limitations and Future Work}
\label{sec:limit}

This work is a preview: it asks whether the real-time video understanding paradigm on a cross-attention backbone is feasible and effective, not whether it reaches state of the art. The evidence supports feasibility---competitive offline video and multimodal understanding, significantly faster inference, and real-time specialization at almost no cost to offline ability---but precisely as a preview, many dimensions remain preliminary.
\subsection{Limitations}
\label{subsec:limitations}

\begin{enumerate}[label=\textbf{(\arabic*)}, leftmargin=1.9em, itemsep=4pt, topsep=2pt]
  \item \textbf{The most critical gap: no quantitative evaluation of real-time understanding.} Our real-time capability is shown only qualitatively. As noted earlier, a real-time evaluation must \emph{jointly} measure accuracy and timeliness, yet decision-level latency---the wait between an event ending and the model deciding to reply---has no standard benchmark today, and accuracy alone can be inflated by ``stalling the answer''. How timely the revisions are and how appropriate the silences are therefore cannot yet be quantified.

  \item \textbf{A gap to SOTA on general benchmarks remains.} The model trails Qwen2.5-VL at a comparable (in fact smaller) scale, especially on OCR, fine-grained perception, and parts of video QA. We attribute this to data and scale, not the cross-attention architecture---we lead on reasoning, spatial, and temporal benchmarks---but the gap is real.

  \item \textbf{Data engineering is preliminary.} Basic understanding and instruction data largely reuse open-source resources; the real-time synthetic data we contribute is limited in scale (836\,K), its synthesis pipeline has not been fully refined, and it has not been open-sourced. Data scale and diversity is the most direct bottleneck today.

  \item \textbf{Ablations are limited.} Many design choices are simple attempts or defaults without systematic controls: the silence threshold ($\tau{=}0.6$), the pooling stride and method, the layer placement of cross-attention, and the real-time-to-offline mixing ratio are not rigorously swept. With validating the paradigm as our focus, we do not attribute individual components rigorously---also why some counter-intuitive results (e.g., the real-time vs.\ offline divergence on SEEDBench and V$^*$) lack a controlled explanation.

  \item \textbf{Limited scale.} Parameters (11B), context length, and training data volume are all limited; the training pipeline is geared toward ``architectural validation'' and does not yet incorporate the mature large-scale efficient training (e.g., 3D parallelism) that further scaling would call for.

  \item \textbf{No RL.} The entire pipeline is next-token SFT. But ``when to speak, when to stay silent, and when to revise'' is at heart a decision involving temporal trade-offs, and SFT can only imitate the pace baked into the synthetic data---it cannot explicitly optimize the ``timely and accurate'' objective.
\end{enumerate}

\subsection{Future work}
\label{subsec:future}

The improvement directions corresponding to the limitations above are, in order:

\begin{enumerate}[label=\textbf{(\arabic*)}, leftmargin=1.9em, itemsep=4pt, topsep=2pt]
  \item \textbf{Establish a decision-level latency benchmark (limitation~(1)).} Design a real-time evaluation protocol and data that \emph{jointly} measure accuracy and decision latency and can detect ``stalling the answer'', making real-time capability quantifiable and comparable for the first time---the prerequisite for everything below.

  \item \textbf{Optimize real-time behavior directly via RL (limitations~(1),~(6)).} Given a quantifiable latency signal, train a reply / silence policy by reinforcement learning with a reward $R = \text{acc} - \lambda \cdot \text{delay}$~\citep{schulman2017ppo,deepseekai2025r1}, with $\lambda$ modulating the accuracy--timeliness trade-off---beyond the fixed pace SFT can only imitate.

  \item \textbf{Scale and diversify the data (limitations~(2),~(3)).} Expand the scale and coverage of both basic understanding and real-time synthetic data (more scenes, longer videos, harder along-the-stream rewrites), and refine and open-source the synthesis pipeline.

  \item \textbf{Scale data, parameters, and context, with efficient training (limitations~(2),~(5)).} Scale along all three axes to narrow the general-benchmark gap, and adopt mature distributed training---Megatron-LM-style~\citep{shoeybi2019megatron,narayanan2021megatron3d} tensor / pipeline / data parallelism---for larger-scale pre-training and fine-tuning. We also plan to open-source the complete training code, weights, and configurations.

  \item \textbf{Channel-wise compression.} Exploit the two-channel separation to compress and quantize the visual K/V alone---it dominates the context budget in multi-frame video---leaving the text untouched, cutting the cost of longer videos and higher frame rates, feeding into the context scaling above.
\end{enumerate}

\section{Conclusion}
\label{sec:conclusion}

This work proposes and preliminarily validates one central direction: advancing video understanding from the offline paradigm toward \textbf{real-time interaction}, in which the model perceives new frames while still replying, revises its answer as new evidence appears, and remains silent when there is nothing to say. Its defining constraint---\textbf{perception must not be blocked by generation}---is naturally realized by a two-channel architecture, so we adopt a \textbf{cross-attention backbone} over the prevailing decoder-only design and train it with a data synthesis pipeline that converts dense captions into real-time understanding QA.

The experiments yield three conclusions. First, the model attains \emph{competitive offline video and multimodal understanding}, and remains robust on the spatial and fine-grained temporal reasoning central to real-time use. Second, on a single H200 with 256 frames per video it achieves approximately a $\mathbf{5\times}$ speedup in time to first token and $\mathbf{2.7\times}$ higher decoding throughput despite its larger size---an advantage that follows from the architecture itself, not an engineered serving stack. Third, real-time specialization incurs \emph{negligible degradation} in offline understanding: ``real-time'' can be added as a nearly lossless capability on top of an offline model.

This remains a preview: its goal is to validate the feasibility of the real-time video understanding paradigm and the cross-attention backbone, not to reach state of the art. We acknowledge a real gap to the strongest models---attributable primarily to data and scale rather than the architecture---and identify the paradigm's most pressing open problem: real-time capability still lacks a quantitative benchmark that jointly measures accuracy and timeliness. Taken together, our study of paradigm, architecture, and data outlines a viable path toward real-time video understanding for the open-source community.

% --- Back matter (optional; uncomment to use) ----------------------------
% \section*{Acknowledgments}
% We thank ... for ...

% --- References ---
\clearpage
\bibliographystyle{unsrtnat}
\bibliography{references}

\end{document}